\documentclass{article}
\usepackage{aaai}
\usepackage{amsfonts}
\usepackage{amssymb}
\usepackage{latexsym}

\newcommand{\equivo}{\equiv_{\top}}
\newcommand{\naturalsplus}{\naturals \cup \{0,\infty\}}

\newcommand{\WEG}[1]{}
\newcommand{\ol}[1]{\overline{#1}}

\renewcommand{\leq}{\leqslant}
\renewcommand{\le}{\leqslant}
\renewcommand{\geq}{\geqslant}

\newcommand{\cL}{{\mathcal L}}
\newcommand{\R}{{\mathcal R}}

\newcommand{\condAB}{(B | A)}

\newcommand{\condABi}[1]{(B_{#1} | A_{#1})}

\newcommand{\notA}{\overline{A}}
\newcommand{\AnotB}{A \overline{B}}
\newcommand{\notB}{\overline{B}}
\newcommand{\condL}{\mbox{$(\cL \mid \cL)$}}
\newcommand{\card}{\mbox{\it card}\,}

\newcommand{\rules}{\{\condABi{1}, \ldots, \condABi{n}\}}

\newcommand{\aiplus}[1]{\mathbf{a}^+_{#1}}
\newcommand{\aiminus}[1]{\mathbf{a}^-_{#1}}

\newcommand{\FR}{\cF_{\R}}
\newcommand{\cF}{{\mathcal F}}

\newcommand{\Omegagroup}{\widehat{\Omega}}
\newcommand{\Omegagroupo}{\Omegagroup_0}
\newcommand{\Omegagroupplus}{\Omegagroup_+}
\newcommand{\omegagroup}{\widehat{\omega}}

\newcommand{\sigmaR}{\sigma_{\R}}
\newcommand{\sigmaAB}{\sigma_{\condAB}}
\newcommand{\Kern}{\mbox{\it ker}\;}
\newcommand{\Kerno}{\mbox{\it ker}_0 \;}

\newcommand{\integers}{\mathbb{Z}}
\newcommand{\naturals}{\mathbb{N}}
\newcommand{\rationals}{\mathbb{Q}}

\newcommand{\D}{\displaystyle}
\newcommand{\hem}{\hspace*{2em}}
\newcommand{\Mod}{\mbox{\it Mod}\,}

\newcommand{\kappaiplus}[1]{\kappa_{#1}^+}
\newcommand{\kappaiminus}[1]{\kappa_{#1}^-}

\newcommand{\condCD}{(D | C)}
\newcommand{\notD}{\overline{D}}

\newcommand{\Zstar}{Z^*}
\newcommand{\kappastar}{\kappa^*}

\renewcommand{\perp}{\,\underline{\, \parallel \,}\,}

\newtheorem{theorem}{Theorem}      
\newtheorem{lemma}[theorem]{Lemma}
\newtheorem{corollary}[theorem]{Corollary}

\newenvironment{definition}{\medskip\par\noindent\refstepcounter{theorem}\textbf{Definition \WEG{\arabic{chapter}.}\arabic{theorem}} \rm}{\medskip\par}

\newenvironment{example}{\medskip\par\noindent\refstepcounter{theorem}\textbf{Example \WEG{\arabic{chapter}.}\arabic{theorem}} \rm}{\hspace*{\fill}\rule{2mm}{2mm}\medskip\par}
\newenvironment{proof}{\medskip\noindent \emph{Proof.} \rm}{\hfill
  $\Box$\medskip\par}
\newenvironment{proofsketch}{\medskip\noindent \emph{Sketch of proof.} \rm}{\hfill $\Box$\medskip\par}
\newenvironment{remark}{\medskip\noindent \em {Remark.} \rm}{\par\medskip}
\title{Conditional indifference\\ and conditional preservation}
\author{Gabriele Kern-Isberner\\ FernUniversitaet Hagen, Department of Computer
  Science \\ P.O.\ Box 940, D-58084 Hagen,
  Germany\\ \texttt{gabriele.kern-isberner@fernuni-hagen.de}}

\begin{document}
  \maketitle

\begin{abstract}
\noindent
The idea of preserving
conditional beliefs emerged recently as a new paradigm apt to guide the revision of
epistemic states. Conditionals are substantially different
from propositional beliefs and need specific treatment. In this paper, we
present a new approach to conditionals, capturing particularly well their
dynamic part as revision policies. We thoroughly axiomatize a \emph{principle
  of conditional preservation} as an indifference property with respect to
\emph{conditional structures} of worlds. This principle is developed in a
semi-quantitative setting, so as to reveal its fundamental meaning for
belief revision in quantitative as well as in qualitative frameworks. In fact,
it is shown to cover other proposed approaches to conditional preservation. 
\end{abstract}

\section{Introduction}

Within the last years, the propositional limitations of classical belief
revision have been overcome piece by piece. For instance, Boutilier
\cite{Boutilier94} investigated belief revision within a modal framework, and Williams
\cite{Williams94} proposed \emph{transmutation schemas for knowledge
  systems}. In general, \emph{epistemic states} have moved into the center of interest as
representations of belief states of some individual or
intelligent agent at a given time. Besides  propositional or
first-order facts, reflecting certain knowledge, they may contain assumptions,
preferences among beliefs, and, as a crucial ingredient, conditional
knowledge. They may be represented in
different ways, e.g.\ by gradings of plausibility or disbelief, by making use
of epistemic entrenchment, or as a probability distribution. 

Epistemic states provide an excellent framework to study
\emph{iterative revisions} which are important to build fully dynamic systems\footnote{An interesting approach to iterated revisions of \emph{belief sets} was
  proposed quite recently by Lehmann, Magidor and Schlechta in
  \cite{LehmannMagidorSchlechta99}.}. 
While propositional AGM theory only observes the results
of revisions, considering epistemic states under change allows one to focus on the
mechanisms underlying that change, 
taking conditional beliefs as \emph{revision
  policies} explicitly into account (cf.\
\cite{BoutilierGoldszmidt93,Rott91,DarwichePearl97}). The connection between
epistemic states, $\Psi$, (iterative) revision operators, $*$, and
conditionals, $\condAB$, is established by the
\emph{Ramsey test}
\begin{equation}
\label{eq_ramsey}
\Psi \models \condAB \quad \mbox{iff} \quad \Psi * A \models B.
\end{equation}
Hence revising epistemic states does not only mean to deal with propositional beliefs -- it also requires studying how conditional beliefs are
changed. 

Darwiche and Pearl \cite{DarwichePearl97} rephrased the AGM postulates for
epistemic states. Applying the \emph{minimal change paradigm} of propositional belief
revision in that extended framework, as Goldszmidt and Boutilier did in
\cite{BoutilierGoldszmidt93}, however, may produce unintuitive results \cite{DarwichePearl97}. So, Darwiche and Pearl \cite{DarwichePearl97} advanced
four postulates as a  cautious approach to describe \emph{principles of conditional
  preservation} when revising epistemic states by  \emph{propositional} beliefs. In \cite{Kern-Isberner99a,Kern-Isberner99b}, we 
extended their approach  in considering revisions of
epistemic states by \emph{conditionals}. We proposed a set of axioms
 outlining conditional revisions which are in accordance with some fundamental postulates of
 revisions (like, for instance, \emph{success}) and with propositional AGM theory, and which
 preserve conditional beliefs in observing conditional interactions. 
These interactions were specified by two
 newly introduced 
 relations between conditionals, that of \emph{subconditionality} and that of
 \emph{perpendicularity} (see \cite{Kern-Isberner99b}).

Earlier, in \cite{Kern-Isberner97e}, we defined a \emph{principle of conditional
   preservation} in quite a different, namely probabilistic, framework. This
 principle there was based on the algebraic notion of conditional
   structures, made use of group theoretical means and postulated the
 numerical values of the given probability distribution to follow the
 \emph{conditional structures} of worlds. 

In this paper, we will bring together both approaches to conditional
preservation, the qualitative one and the quantitative one, in the
semi-quantitative framework of ordinal conditional functions. We will rephrase
the probabilistic principle of conditional preservation of
\cite{Kern-Isberner97e} for ordinal conditional functions, and we will show,
that 
this quantitative principle of conditional
preservation implies the corresponding qualitative postulates of
\cite{Kern-Isberner99b}. Actually, the quantitative version is much stronger
than the qualitative one,
dealing with \emph{sets of conditionals} instead of only one revising
conditional,  and handling interactions of
conditionals of arbitrary complexity. Although numerical in nature, the principle is
based on a symbolic representation of conditional influences on worlds, called
\emph{conditional structures}. Numbers (rankings or probabilities) are only
considered as manifestations of complex conditional interactions. The
representation of conditional structures by the aid of group theory
provides a rich methodological framework to study conditionals in belief
revision and nonmonotonic reasoning. 

In the following section, 
we fix 
some notations  and describe the relationship between conditionals and
epistemic states. 
Then we introduce the crucial notion of
conditional structures which the property of conditional indifference is based
upon.
By applying the concept of
indifference to revision functions, we obtain a precise formalization of the
principle of conditional preservation, and we characterize ordinal conditional
functions observing this principle. 
Further on, we compare our approach to Goldszmidt, Morris \& Pearl's
system-Z and system-$\Zstar$ in several examples. 
Finally, we bring together qualitative and
quantitative approaches to the principle of conditional preservation, proving
the formalization given here to be a most fundamental one. A summary and an
outlook  conclude this paper.

\section{Conditionals and epistemic states}
\label{sec_conditionals_epistemic_states}
We consider a propositional language $\cL$ over a finite
alphabet $a,b,c \ldots$. Let $\Omega$ denote the set of possible worlds for $\cL$, i.e.\ $\Omega$ is a  complete set of interpretations of $\cL$. Throughout
this paper, 
we will write $\ol{A}$
instead of $\neg A$, and $AB$ instead of $A \wedge B$, for formulas $A,B \in
\cL$. 

Conditionals $\condAB$ represent statements of the form ``\emph{If A then
  B}'', expressing a relationship between two 
(propositional) formulas $A$, the \emph{antecedent} or \emph{premise}, and $B$,
the \emph{consequent}. 
$\condL$ denotes the set of all conditionals
$\condAB$ with $A,B \in \cL$. A conditional $(B|\top)$ with a tautological
  antecedent is taken to correspond to its (propositional) consequent, $B$. 
$\condCD$ is called a \emph{subconditional} of $\condAB$, written as 
\begin{equation}
\label{eq_subconditional}
\condCD \sqsubseteq \condAB,
\end{equation}
iff $CD \models AB$
  and $C\notD \models \AnotB$. Typically, subconditionals arise by
  strengthening the antecedent of a conditional, e.g.\ $(b|ac)$ is a
  subconditional of $(b|a)$, $(b|ac) \sqsubseteq (b|a)$.

Each possible world $\omega \in \Omega$ either
confirms $\condAB$, in case that $\omega \models AB$, or refutes it, if
  $\omega \models \AnotB$, or does not even satisfy its premise, $\omega
  \not\models A$, and so is of no relevance for it.  Although conditionals are evaluated with
respect to worlds, they cannot really be accepted (as entities) in single, isolated worlds. To validate conditionals, we
need  richer epistemic structures than  plain propositional interpretations, at least to compare different
worlds with regard to their relevance for a conditional (see, for example,
\cite{Nute80,Boutilier94,DarwichePearl97}). Epistemic states
as representations of cognitive states of intelligent agents provide an
adequate framework for conditionals. 

An epistemic notion that turned out to be of great importance both for
conditionals and epistemic states, in particular in the context of  belief revision, is that of \emph{plausibility}: conditionals
are supposed to represent plausible conclusions, and plausibility relations
on formulas or worlds, respectively, guide AGM-revisions of belief sets and of epistemic states \cite{Nute80,KatsunoMendelzon91a,DarwichePearl97}.

As Spohn
\cite{Spohn88} emphasized, however, it is not enough to consider the
qualitative ordering of propositions according to their plausibility -- also 
relative distances between degrees of plausibility should be taken into account. So
he introduced ordinal conditional functions $\kappa$ (\emph{OCF's, ranking functions})
\cite{Spohn88}  from worlds 
to ordinals such that some worlds are mapped to the minimal element $0$.
Here, we will simply assume that OCF's are functions $\kappa: \Omega \to
\naturals \cup \{0,\infty\}$ from the set of worlds 
to the natural numbers, extended by $0$ and $\infty$. 
They specify non-negative integers as
degrees of plausibility -- or, more precisely, as degrees of \emph{disbelief}
-- for worlds.  The smaller $\kappa(\omega)$ is, the more plausible
the world $\omega$ appears, and what is believed (for certain) in the epistemic
state represented by $\kappa$ is described precisely by the set 
$\Mod(\kappa) := \{ \omega \in \Omega \mid \kappa(\omega)
 = 0 \}$. 
For  propositional
formulas $A,B \in \cL$, we set $\kappa(A) = \min\{\kappa(\omega) \mid \omega
\models A\}$, so that $\kappa(A \vee B) = \min\{\kappa(A), \kappa(B) \}$.
In particular, $0 = \min
\{\kappa(A), \kappa(\notA)\}$, so that at least one of $A$ or $\notA$ is
considered mostly plausible.
A proposition $A$ is believed iff $\kappa(\notA)
>0$, which is denoted by $\kappa \models A$.
A conditional $\condAB \in \condL$ may be assigned a degree of plausibility via
%
\WEG{
\[
\kappa(B |A) = \kappa(AB) - \kappa(A) = \left \{
\begin{array}{l@{\; : \;}l}
0 & \kappa(AB) \leq \kappa(\AnotB)\\
\kappa(AB) - \kappa(\AnotB) & \kappa(AB) \geq \kappa(\AnotB)
\end{array} \right.
\]
}
%
$\kappa(B |A) = \kappa(AB) - \kappa(A)$.
Each OCF $\kappa$ induces a (propositional) AGM-revision operator
$*$ by setting $\Mod(\kappa * A) = \min\nolimits_{\kappa}(\Mod(A))$ (see \cite{DarwichePearl97}). The Ramsey
test (\ref{eq_ramsey}) then reads $\kappa \models \condAB$ iff $\kappa * A
\models B$. This is in accordance with the
plausibility relation imposed by $\kappa$, as the following lemma shows:
\begin{lemma}
\label{revepst_lemma_conditional_equivalences}
Let $\condAB$ be a conditional in $\condL$, let $\kappa$ be an ordinal
conditional function. Then $\kappa  \models \condAB$ (by applying the Ramsey
test) iff  $\kappa(AB) < \kappa(\AnotB)$.
\end{lemma}
So $\kappa$ accepts a conditional (via the Ramsey test) iff $AB$ is more
plausible than $\AnotB$. The proof of this lemma is immediate.

\section{Conditional structures}
\label{sec_conditional_structures}

By observing the behavior of worlds with respect to it, each conditional $\condAB$ can be considered as a generalized (namely three-valued)
  indicator function on worlds: 
\begin{equation}
\label{conditionals_eq_conditional}
\condAB(\omega)= \left\{ 
\begin{array}{r@{\quad : \quad}l} 1 & \omega \models AB \\ 0 & \omega \models
  \AnotB \\ u & \omega \models  \notA
\end{array} \right.
\end{equation}                                             
where $u$ stands for {\em undefined} \cite{deFinetti74,calabrese91}).
Intuitively, 
incorporating a conditional as a plausible conclusion in an epistemic state means to make -- at least
some -- worlds confirming the conditional more plausible than the worlds
refuting it. In this sense, conditionals to be learned have effects on
possible worlds (more exactly, on their degrees of plausibility), shifting them
appropriately to establish the intended plausible relationship.
(\ref{conditionals_eq_conditional}) then provides a classification of worlds
for achieving this: On confirming worlds $\omega \models AB$, i.e.\
$\condAB(\omega) =1$, $\condAB$ possibly has a positive effect, while on
refuting worlds $\omega \models \AnotB$, $\condAB$ possibly has a negative
effect; the effects on worlds $\omega$ with $\condAB(\omega) = u$ is
unclear. Which worlds will actually be shifted depends on the chosen revision
procedure -- for the conditional, all worlds in either of the partitioning
sets are indistinguishable. 

When we consider (finite) sets of conditionals $\R = \rules \subseteq \condL$, we have to
modify the representation (\ref{conditionals_eq_conditional}) appropriately to
identify the effect of each conditional in $\R$ on worlds in
$\Omega$. This leads to introducing the functions $\sigma_i =
\sigma_{\condABi{i}}$ below (see (\ref{conditionals_eq_sigmai})) which
generalize (\ref{conditionals_eq_conditional}) by replacing the numbers $0$
and $1$ by abstract symbols. Moreover, we will make use of a group structure to represent the
joint impact of conditionals on worlds.

To each conditional $\condABi{i}$ in $\R$ we associate two
symbols  $\aiplus{i},\aiminus{i}$. Let 
\[
\FR =\langle \aiplus{1},\aiminus{1}, \ldots,\aiplus{n},\aiminus{n} \rangle
\]
be the free abelian group with generators $\aiplus{1},\aiminus{1}, \ldots,
\aiplus{n},\aiminus{n}$, i.e.\ $\FR$  consists of all elements of the
form $(\aiplus{1})^{r_{1}} (\aiminus{1})^{s_{1}} \ldots (\aiplus{n})^{r_{n}} (\aiminus{n})^{s_{n}}$
with integers $r_{i},s_{i} \in \integers$ (the ring of integers). Each
element of $\FR$ can be identified by its exponents, so that $\FR$ is isomorphic to
$\integers^{2n}$ \cite{LyndonSchupp77}. The commutativity of $\FR$ corresponds to the fact
that the conditionals in $\R$ shall be effective simultaneously, without
assuming any order of application. So our way of dealing with conditionals is
a symmetric, homogeneous one -- we do not need (user-defined) priorities among
conditionals. Note that, although we will 
speak of \emph{multiplication} and \emph{products} in $\FR$, the generators of
$\FR$ are merely juxtaposed, like words.

For each $i,1\le i \le n$, we define a
function $\sigma_{i} : \Omega \to \FR$ by setting
\begin{equation}
\label{conditionals_eq_sigmai}
\sigma_{i}(\omega) = \left\{ 
\begin{array}{r@{\quad \mbox{if} \quad}l} \aiplus{i} & \condABi{i}(\omega)=1 \\ \aiminus{i} & \condABi{i}(\omega)=0 \\ 1 & \condABi{i}(\omega)=u
\end{array} \right.
\end{equation}
$\sigma_{i}(\omega)$ represents the manner in which the conditional
$\condABi{i}$ applies to the possible world $\omega$.   The neutral element $1$ of
$\FR$  corresponds to the non-applicability of $\condABi{i}$ in case that
the antecedent $A_{i}$  is not satisfied. The function $\sigmaR : \Omega \to
\FR$,
\[
\sigmaR(\omega) = \prod_{1 \le i \le n}\sigma_{i}(\omega)=\prod_{1 \le i \le
  n \atop \omega \models A_i B_i }
  \aiplus{i} \prod_{1 \le i \le n \atop \omega \models A_{i}\notB_{i}} \aiminus{i}
\]
describes the all-over effect of $\R$ on $\omega$. $\sigmaR(\omega)$ is called
 \emph{(a representation of) the conditional structure of $\omega$ with
 respect to $\R$}\index{conditional structure!of $\omega$, representation
 of}. 
For each world $\omega$, 
$\sigmaR(\omega)$  contains
at most one of each $\aiplus{i}$ or $\aiminus{i}$, but never both of them because each
conditional applies to $\omega$  in a well-defined way. The next lemma (which
 is easy to prove) shows
 that this property characterizes conditional structure functions:
\begin{lemma}
\label{cond_lemma_sigmarepresentation}
Let $\sigma : \Omega \to \cF$ be a map from the set of worlds $\Omega$ to the
free abelian group $\cF = \langle \aiplus{1},\aiminus{1},
\ldots,\aiplus{n},\aiminus{n} \rangle$ generated by $\aiplus{1},\aiminus{1},
\ldots,\aiplus{n},\aiminus{n}$, such that $\sigma(\omega)$  contains
at most one of each $\aiplus{i}$ or $\aiminus{i}$, for each world $\omega \in \Omega$. Then there is a set of
conditionals $\R$ with $\card(\R) \leq n$ such that $\sigma = \sigmaR$. 
\end{lemma}
\begin{example}
\label{cond_ex_sigmaR}
Let $\R = \{(c|a), (c| b)\}$, where $a,b,c$ are atoms, and let $\FR =
\langle \aiplus{1}, \aiminus{1}, \aiminus{2}, \aiminus{2} \rangle$. We
associate $\mathbf{a}^{\pm}_{1}$ with the first conditional, $(c|a)$, and
$\mathbf{a}^{\pm}_{2}$ with the second one, $(c| b)$. The
following table shows the values of the function $\sigmaR$ on worlds $\omega
\in \Omega$:
\begin{displaymath}
\begin{array}{|l|l||l|l|}
\hline
\omega & \sigmaR(\omega) & \omega & \sigmaR(\omega)\\ \hline
abc  \rule[0mm]{0mm}{5mm} & \aiplus{1} \aiplus{2} & \ol{a} bc & \aiplus{2}\\
ab \ol{c} & \aiminus{1} \aiminus{2} & \ol{a} b \ol{c} & \aiminus{2}\\
a \ol{b} c  & \aiplus{1}  & \ol{a}  \ol{b} c & 1\\
a \ol{b} \ol{c} & \aiminus{1} & \ol{a} \ol{b}  \ol{c} & 1\\
\hline
\end{array}
\end{displaymath}
$abc$ confirms both conditionals, so its conditional structure is represented by
$\aiplus{1} \aiplus{2}$. This corresponds to the product (in $\FR$) of the
conditional structures of the worlds $\ol{a} bc$ and $a \ol{b} c$. Two worlds, namely $\ol{a}  \ol{b} c$ and $\ol{a} \ol{b}  \ol{c}$, are
not affected at all by the conditionals in $\R$. 
\end{example}
%
%
\WEG{
\begin{example}
\label{cond_ex_sigmaR}
Let $\R = \{(b|a), (b|ac)\}$, where $a,b,c$ are atoms, and let $\FR =
\langle \aiplus{1}, \aiminus{1}, \aiminus{2}, \aiminus{2} \rangle$. We
associate $\mathbf{a}^{\pm}_{1}$ with the first conditional, $(b|a)$, and
$\mathbf{a}^{\pm}_{2}$ with the second one, $(b| ac)$. The
following table shows the values of the function $\sigmaR$ on worlds $\omega
\in \Omega$: 
\begin{displaymath}
\begin{array}{|l|l||l|l|}
\hline
\omega & \sigmaR(\omega) & \omega & \sigmaR(\omega)\\ \hline
abc  \rule[0mm]{0mm}{5mm} & \aiplus{1} \aiplus{2} & \ol{a} bc & 1\\
ab \ol{c} & \aiplus{1}  & \ol{a} b \ol{c} & 1\\
a \ol{b} c  & \aiminus{1} \aiminus{2}  & \ol{a}  \ol{b} c & 1\\
a \ol{b} \ol{c} & \aiminus{1} & \ol{a} \ol{b}  \ol{c} & 1\\
\hline
\end{array}
\end{displaymath}
$abc$ confirms both conditionals, so its conditional structure is represented by
$\aiplus{1} \aiplus{2}$. Four worlds, namely $\ol{a} bc, \ol{a} b \ol{c},
\ol{a}  \ol{b} c, \ol{a} \ol{b}  \ol{c}$, are
not affected at all by the conditionals in $\R$. 
\end{example}
}
%
The logical structure of antecedents and consequents of the conditionals in
$\R$ does not really matter, nor do logical relationships between the
conditionals. All that we need is a conditional's partitioning property on the
set of worlds (cf.\ (\ref{conditionals_eq_conditional}) and
(\ref{conditionals_eq_sigmai})). $\sigmaR$ labels each world
appropriately and allows us to compare different worlds with respect to the
impact the conditionals in $\R$ exert on them. 
The following example illustrates that also multiple copies of worlds may be
necessary to relate conditional structures:
\begin{example}
\label{cond_ex_sigmaR2}
Consider the set $\R = \{(d|a), (d|b), (d|c)\}$ of conditionals using the
atoms $a,b,c,d$. Let $\mathbf{a}^{\pm}_{1}, \mathbf{a}^{\pm}_{2},
\mathbf{a}^{\pm}_{3}$ be the group generators associated with $(d|a), (d|b)$,
$(d|c)$, respectively. Then we have
\begin{eqnarray*}
\sigmaR(ab\ol{c}d) \sigmaR(a\ol{b}cd) \sigmaR(\ol{a}bcd)
&=&(\aiplus{1}\aiplus{2}) (\aiplus{1}\aiplus{3}) (\aiplus{2}\aiplus{3}) \\
= (\aiplus{1})^2 (\aiplus{2})^2 (\aiplus{3})^2 
&=& (\aiplus{1} \aiplus{2} \aiplus{3})^2 \\
&=& \sigmaR(abcd)^2.
\end{eqnarray*}
Here two copies of $abcd$, or of its structure, respectively, are necessary to
match the product of the conditional structures of $ab\ol{c}d, a\ol{b}cd$ and $\ol{a}bcd$.
\end{example}

To compare worlds adequately with respect to their conditional structures, we
take the worlds $\omega \in \Omega$ as formal generators of the free abelian
group 
%
\[
\label{cond_def_Omegagroup} \index{$\Omegagroup$}
\Omegagroup := \langle \omega \mid \omega \in \Omega \rangle
\]
$\Omegagroup$ consists of all products $\omegagroup = {\omega_1}^{r_1} \ldots
{\omega_m}^{r_m}$, with $\omega_1, \ldots, \omega_m \in
\Omega$,  and  $r_1, \ldots r_m$ integers.
Introducing such a ``multiplication between worlds'' is nothing but a
technical means to comply with the multiplicative structure the effects of
conditionals impose on worlds. As in $\FR$, multiplication in $\Omegagroup$ actually means
juxtaposition. In \cite{Kern-Isberner97e}, where we first developed these
ideas, we considered multi-sets of worlds (corresponding to elements in
$\Omegagroup$ with only positive exponents) and calculated the conditional
structure of such a multi-set as the \emph{conditional weight} it is
carrying. Making use of arbitrary elements of $\Omegagroup$ as group elements, however, provides a much more convenient and
elegant framework to deal with conditional structures. 
We will usually write 
$\D\frac{\omega_1}{\omega_2}$ instead of $\omega_1 \omega_2^{-1}$. 

Now $\sigmaR$ may be extended to $\Omegagroup$ 
in a straightforward manner by setting
\[
\sigmaR (\omegagroup) = \sigmaR(\omega_1)^{r_1} \ldots \sigmaR(\omega_m)^{r_m},
\]
yielding a \emph{homomorphism of groups} $\sigmaR: \Omegagroup \to \FR$. For  $\omegagroup = {\omega_1}^{r_1} \ldots {\omega_m}^{r_m} \in
\Omegagroup$, we obtain 
\begin{eqnarray*}
\label{conditionals_eq_condstruc1}
&& \sigmaR({\omega_1}^{r_1} \ldots {\omega_m}^{r_m}) =\\ 
&& \hspace*{1em} \prod_{1 \leq i \leq
  n}(\aiplus{i})^{\sum_{k: \sigma_i(\omega_k) = \aiplus{i}}r_k} \prod_{1 \leq i \leq
  n}(\aiminus{i})^{\sum_{k: \sigma_i(\omega_k) = \aiminus{i}}r_k},
\end{eqnarray*}
as a representation of its conditional structure. 
\WEG{
is represented by a group element which is a product of the generators
$\aiplus{i},\aiminus{i}$ of $F_{\R}$, with each $\aiplus{i}$ occurring with exponent
$\sum_{k:\sigma_{i}(\omega_{k})=\aiplus{i}}r_{k} = \sum_{k: \omega_{k} \models
  A_i B_i}r_{k}$, and each $\aiminus{i}$ occurring with exponent
$\sum_{k:\sigma_{i}(\omega_{k})=\aiminus{i}}r_{k} = \sum_{k: \omega_{k}
  \models A_i \notB_i}r_{k}$ (note that each of the sums may be empty in which
  case  the corresponding conditional cannot be applied to any of the
  worlds occurring in $\omegagroup$). 
}
The exponent of
  $\aiplus{i}$ in $\sigmaR(\omegagroup)$ indicates the number of worlds
  in $\omegagroup$ which confirm the conditional $\condABi{i}$, each world
being counted with its multiplicity, and in the same way, the exponent of
$\aiminus{i}$ indicates the number of worlds that are in conflict with
$\condABi{i}$.  

By investigating suitable elements of $\Omegagroup$, it is possible to isolate the (positive or negative) net impacts of conditionals in $\R$ , as the
following example illustrates:

\begin{example}(continued)
In Example \ref{cond_ex_sigmaR} above, we have 
\[
\sigmaR(\frac{abc}{\ol{a} bc})  = \frac{\aiplus{1} \aiplus{2}}{\aiplus{2}}
                                =  \aiplus{1}
\]
So $\D\frac{abc}{\ol{a} bc}$ reveals the positive net impact of the conditional $(c|a)$
within $\R$, symbolized by $\aiplus{1}$.

Similarly,
in Example \ref{cond_ex_sigmaR2}, the element
$\D\frac{ab\ol{c}\ol{d} \cdot \ol{a}\ol{b}c\ol{d}}{a\ol{b}c\ol{d}}$ isolates
the negative net impact of the second conditional, $(d|b)$:\\
\hem 
$\sigmaR\left(\D\frac{ab\ol{c}\ol{d} \cdot \ol{a}\ol{b}c\ol{d}}{a\ol{b}c\ol{d}}\right) = 
\D\frac{\aiminus{1}\aiminus{2} \cdot \aiminus{3}}{\aiminus{1}\aiminus{3}}
= \aiminus{2}$.
\end{example} 

The following example is taken from \cite[p.\ 68f]{GoldszmidtPearl96}:
\begin{example}
\label{revepst_ex_penguin}
Consider the set $\R$ consisting of the following conditionals:
\[
\begin{array}{l@{\; : \;}l@{\quad}l}
r_1 &  (f |b) & \mbox{\emph{Birds fly.}} \\
r_2 &  (b |p) & \mbox{\emph{Penguins are birds.}}\\
r_3 &  (\ol{f} | p) & \mbox{\emph{Penguins do not fly.}} \\
r_4 &  (w |b) & \mbox{\emph{Birds have wings.}}\\
r_5 &  (a |f) & \mbox{\emph{Animals that fly are airborne.}} 
\end{array}
\]
\begin{table}[h]
\begin{displaymath}
\begin{array}{|l|l||l|l|}
\hline
\omega & \sigmaR(\omega) &  \omega & \sigmaR(\omega)  \\
\hline
\rule{0mm}{4mm} pbfwa & \aiplus{1}\aiplus{2}\aiminus{3}\aiplus{4}\aiplus{5} 
& \ol{p}bfwa & \aiplus{1}\aiplus{4}\aiplus{5} \\
pbfw\ol{a} & \aiplus{1}\aiplus{2}\aiminus{3}\aiplus{4}\aiminus{5} 
& \ol{p}bfw\ol{a} & \aiplus{1}\aiplus{4}\aiminus{5} \\
pbf\ol{w}a & \aiplus{1}\aiplus{2}\aiminus{3}\aiminus{4}\aiplus{5}
& \ol{p}bf\ol{w}a & \aiplus{1}\aiminus{4}\aiplus{5} \\
pbf\ol{w}\,\ol{a} & \aiplus{1}\aiplus{2}\aiminus{3}\aiminus{4}\aiminus{5}
& \ol{p}bf\ol{w}\,\ol{a} & \aiplus{1}\aiminus{4}\aiminus{5} \\

\rule{0mm}{4mm} pb\ol{f}wa & \aiminus{1}\aiplus{2}\aiplus{3}\aiplus{4}
& \ol{p}b\ol{f}wa & \aiminus{1}\aiplus{4} \\
pb\ol{f}w\ol{a} & \aiminus{1}\aiplus{2}\aiplus{3}\aiplus{4}
& \ol{p}b\ol{f}w\ol{a} & \aiminus{1}\aiplus{4} \\
pb\ol{f}\ol{w}a & \aiminus{1}\aiplus{2}\aiplus{3}\aiminus{4}
& \ol{p}b\ol{f}\ol{w}a & \aiminus{1}\aiminus{4}\\
pb\ol{f}\ol{w}\,\ol{a} & \aiminus{1}\aiplus{2}\aiplus{3}\aiminus{4}
& \ol{p}b\ol{f}\ol{w}\,\ol{a} & \aiminus{1}\aiminus{4} \\

\rule{0mm}{4mm} p\ol{b}fwa & \aiminus{2}\aiminus{3}\aiplus{5}
& \ol{p}\ol{b}fwa & \aiplus{5} \\
p\ol{b}fw\ol{a} & \aiminus{2}\aiminus{3}\aiminus{5}
& \ol{p}\ol{b}fw\ol{a} &  \aiminus{5}\\
p\ol{b}f\ol{w}a & \aiminus{2}\aiminus{3}\aiplus{5}
& \ol{p}\ol{b}f\ol{w}a & \aiplus{5} \\
p\ol{b}f\ol{w}\,\ol{a} & \aiminus{2}\aiminus{3}\aiminus{5}
& \ol{p}\ol{b}f\ol{w}\,\ol{a} & \aiminus{5} \\

\rule{0mm}{4mm} p\ol{b}\,\ol{f}wa & \aiminus{2}\aiplus{3}
& \ol{p}\ol{b}\,\ol{f}wa & 1 \\
p\ol{b}\,\ol{f}w\ol{a} & \aiminus{2}\aiplus{3} 
& \ol{p}\ol{b}\,\ol{f}w\ol{a} & 1 \\
p\ol{b}\,\ol{f}\ol{w}a & \aiminus{2}\aiplus{3} 
& \ol{p}\ol{b}\,\ol{f}\ol{w}a & 1 \\
p\ol{b}\,\ol{f}\ol{w}\,\ol{a} & \aiminus{2}\aiplus{3} 
& \ol{p}\ol{b}\,\ol{f}\ol{w}\,\ol{a} & 1 \\
\hline
\end{array}
\end{displaymath}
\caption{\label{revepst_figure_penguin} Conditional structures for  Example \ref{revepst_ex_penguin}}
\end{table}
In Table \ref{revepst_figure_penguin}, we  list the conditional structures
of all possible worlds; this table will be helpful in the sequel. 
\end{example}

Having the
same conditional structure defines an equivalence relation $\equiv_{\R}$  on
$\Omegagroup$:
\begin{equation}
\label{eq_equivalence_relation}
\omegagroup_1 \equiv_{\R} \omegagroup_2 \quad \mbox{iff} \quad
\sigmaR(\omegagroup_1) = \sigmaR(\omegagroup_2).
\end{equation} 
Those elements of $\Omegagroup$ that are balanced with respect to the effects
of conditionals in $\R$ are contained in the \emph{kernel of $\sigmaR$}, 
$\Kern \sigmaR = \{ \omegagroup \in \Omegagroup \mid \sigmaR(\omegagroup) =
1\}$.
$\Kern \sigmaR$ does not depend on the chosen representation of
conditional structures by symbols
in $\FR$ and thus, it is an invariant of $\R$ \cite{Kern-Isberner99d}. 

Often, besides the conditionals explicitly given in $\R$, implicit normalizing
constraints have to be taken into account, like, e.g.\, $\kappa(\top)=0$ for
ordinal conditional functions. This can be
achieved by focusing on equivalence with repect to $\sigma_{\top}$. 
Since $\sigma_{\top}$ simply counts the generators occurring in
$\omegagroup$, 
%
\WEG{  
\[
\Omegagroupo 
= \left\{ \omegagroup = {\omega_1}^{r_1} \cdot \ldots \cdot
{\omega_m}^{r_m} \in \Omegagroup \mid \sum_{j = 1}^{m} r_j = 0\right\}.
\]
}
%
two elements $\omegagroup_1 = \omega_1^{r_1} \ldots \omega_m^{r_m},\, 
\omegagroup_2 = \nu_1^{s_1} \ldots \nu_p^{s_p} \in \Omegagroup$ are $\sigma_{\top}$-equivalent,  $\omegagroup_1 \equivo \omegagroup_2$,
iff $\sum_{1 \leq j \leq m} r_j = \sum_{1 \leq k \leq p} s_k$.
This means, $\omegagroup_1 \equivo \omegagroup_2$  iff they both are a (cancelled) product of the same number of
generators, each generator being counted with its corresponding exponent.
\WEG{
\noindent
Let
\begin{displaymath}
\Kerno \sigmaR := \Kern \sigmaR \cap \Omegagroupo
\end{displaymath}
be the part of $\Kern \sigmaR$ which is included in $\Omegagroupo$. $\Kerno
\sigmaR$ is less expressive than $\Kern \sigmaR$, for instance,  it does
not contain all $\omega \in \Omega$ with $\sigmaR(\omega) = 1$. But $\Kerno
\sigmaR$ concentrates
on considering ratios as essential entities to reveal the influences of
conditionals. The following lemma shows that $\Kerno \sigmaR$ differs from
$\Kern \sigmaR$
only  by taking the
conditional tautology $(\top \mid \top)$  into regard:

\begin{lemma}
\label{cond_lemma_kerno}
$\Omegagroupo =  \Kern \sigma_{(\top | \top)}$, and $\Kerno \sigmaR = \Kern \sigma_{\R \cup \{(\top \mid \top)\}}$.
\end{lemma}
}

\section{Conditional indifference}
\label{sec_conditional_indifference}
To study conditional interactions, we now focus on the behavior of OCF's $\kappa: \Omega \to \naturalsplus$ with
respect to the multiplication 
in $\Omegagroup$.
Each such function may be extended to a
homomorphism, 
$\kappa: \Omegagroupplus \to (\integers,+)$, by setting
\[
\kappa({\omega_1}^{r_1} \cdot \ldots \cdot {\omega_m}^{r_m}) = r_1 \kappa(\omega_1)
+ \ldots + r_m \kappa(\omega_m),
\]
where $\Omegagroupplus$\index{$\Omegagroupplus$} is the subgroup of $\Omegagroup$ generated by the set
$\Omega_+ := \{\omega \in \Omega \mid \kappa(\omega) \neq \infty\}$. 
This allows us to analyze numerical relationships holding between different
$\kappa(\omega)$. Thereby, it will be possible  to elaborate the conditionals whose
structures $\kappa$ follows, that means, to determine sets of conditionals $\R \subseteq \condL$
with respect to which $\kappa$ is indifferent:

\begin{definition}
\label{cond_def_conditional_indifference}
\index{conditional indifference!strict conditional indifference with respect
  to $\R$}
\index{conditional indifference!weak conditional indifference with respect to $\R$}
Suppose $\kappa: \Omega \to \naturalsplus$ is an OCF, and $\R \subseteq
\condL$ is a set of conditionals such that $\kappa(A) \neq \infty$ for all
$\condAB \in \R$. 
$\kappa$  is \emph{indifferent with
  respect to } $\R$ iff the following two conditions hold:
\begin{enumerate}
\item[(i)] If $\kappa(\omega) = \infty$ then there is $\condAB \in \R$ such that
  $\sigmaAB(\omega) \neq 1$ and $\kappa(\omega') = \infty$ for all $\omega'$ with
  $\sigmaAB(\omega') = \sigmaAB(\omega)$. 
\item[(ii)] $\kappa(\omegagroup_1) = \kappa(\omegagroup_2) \quad \mbox{whenever } \;
\sigmaR(\omegagroup_1) = \sigmaR(\omegagroup_2)$
for  $\omegagroup_1 \equivo \omegagroup_2 \in \Omegagroupplus$.
\end{enumerate} 
\end{definition}
If $\kappa$ is indifferent with respect to $\R \subseteq \condL$, then it does not
distinguish between different elements $\omegagroup_1, \omegagroup_2$ with the
same conditional structure with respect to $\R$. Normalizing constraints are taken
 into account by observing $\equivo$-equivalence. Conversely, any deviation
$\kappa(\omegagroup) \neq 0$ can be explained by the conditionals in $\R$ acting
on $\omegagroup$ in
a non-balanced way. 
Condition (i) in Definition
\ref{cond_def_conditional_indifference} is necessary to deal with worlds
$\omega \notin \Omega_+$.
Conditional indifference, as defined in Definition
\ref{cond_def_conditional_indifference}, captures interactions of conditionals
of arbitrary depth by making use of the homomorphism induced by $\kappa$. It
also respects, however, indifference on the superficial level of the function
$\kappa$ itself:

\begin{lemma}
\label{cond_lemma_conditional_indifference1}
If the ordinal conditional  function $\kappa$ is  indifferent with respect to $\R$,
then $\sigmaR(\omega_1) = \sigmaR(\omega_2)$ implies $\kappa(\omega_1) = \kappa(\omega_2)$ for all worlds $\omega_1, \omega_2 \in \Omega$. 
\end{lemma}

The next theorem gives a simple criteria to check
conditional indifference with ordinal conditional
functions. 
Moreover, it provides an intelligible schema to construct conditional indifferent functions.

\begin{theorem}
\label{cond_cor_char_weak_indifference_ocf}
An OCF $\kappa$ is  indifferent
with respect to a set $\R = \{\condABi{1}, \ldots, \condABi{n}\}$ of conditionals iff $\kappa(A_i) \neq \infty$ for all \mbox{$i, 1 \leq i \leq n$}, and  there
are rational numbers $\kappa_0, \kappaiplus{i}, \kappaiminus{i} \in \rationals$, \mbox{$1 \leq i
\leq n$},  such that for all $\omega \in \Omega$,
\begin{equation}
\label{cond_eq_char_weak_indifference_ocf}
\kappa(\omega) = \kappa_0 + \sum_{1 \leq i \leq n \atop \omega \models A_i B_i}
\kappaiplus{i} + \sum_{1 \leq i \leq n \atop \omega \models A_i \ol{B_i}}
\kappaiminus{i} 
\end{equation}
\end{theorem}
\begin{proofsketch}
According to Lemma \ref{cond_lemma_conditional_indifference1}, the equivalence relation (\ref{eq_equivalence_relation}) provides a rough
classification of the worlds in $\Omega$ with respect to the conditionals in
$\R$. Obtaining a representation of the form
(\ref{cond_eq_char_weak_indifference_ocf}) then amounts to 
checking the solvability of a linear
equational system. The proof of this theorem is very similar to the proof of the analogous theorem
for probabilistic representation of knowledge given in \cite{Kern-Isberner97e}.
\end{proofsketch}
\WEG{
Note that the concept of conditional indifference is a  structural notion,
using the numerical values of  an ordinal  conditional function, $\kappa$, as
manifestations of  conditional structures imposed by a set $\R$. We do not
postulate, however, that the conditionals in $\R$ are satisfied by $V$. This
adoption of $\R$ will be dealt with in the following section.  
}

\section{The principle of conditional preservation}
\label{sec_principle_conditional_preservation}

Minimality of change is a crucial paradigm for belief revision, and a ``principle of
conditional preservation'' is to realize this idea of minimality when
conditionals are involved in change. Minimizing absolutely the changes in
conditional beliefs, as in \cite{BoutilierGoldszmidt93}, is an important
proposal to this aim, but it  does not always lead
to intuitive results \cite{DarwichePearl97}. The idea we will
develop here rather aims at \emph{preserving the conditional structure of
  knowledge} within an epistemic state which we assume to be represented by an
OCF $\kappa$. 
\WEG{
The notion of a conditional structure with respect to a set $\R$ of
conditionals was defined in Section \ref{cond_sec_conditional_structures}, and
in Section \ref{cond_sec_conditional_indifference}, }

We just explained what it means
 for an OCF $\kappa$ to follow the structure imposed by $\R$ on the set of worlds by
 introducing the notion of conditional indifference (cf.\ Definition
 \ref{cond_def_conditional_indifference}). 
Pursuing this approach further in the framework of belief revision, a revision
of $\kappa$ by simultaneously incorporating the conditionals in $\R$, $\kappa^* = \kappa *
\R$, can be said to
preserve the conditional structure of $\kappa$ with respect to $\R$ if the
\emph{relative change function} $\kappa^* - \kappa$  is indifferent with respect
to $\R$\footnote{Why just \mbox{$\kappa^* -
 \kappa$}? First, it is more accurate  than, e.g., $\max\{0,\kappa^* -
 \kappa\}$, in the sense of taking differences in degrees of plausibility seriously. Second,
 it makes use of the conditional ``$-$'', considering revision as a
 generalized conditional operation; for more details, see \cite{Kern-Isberner99d}.}. Taking into regard the worlds $\omega$ with $\kappa(\omega) = \infty$
appropriately, this gives rise to the following definitions:

\begin{definition}
\label{revepst_def_principle_conditional_preservation}
\index{conditional preservation!principle of conditional preservation}
\index{consistency!V-consistency}
Let $\kappa$ be an OCF, 
and let $\R$ be a
finite set of conditionals. Let $\kappa^* = \kappa * \R$ denote the
result of revising $\kappa$ by $\R$. Presuppose further\footnote{Note that
  success $\kappastar \models \R$ is not compulsory for conditional
  indifference and conditional 
  preservation. We  only presuppose $\kappa^*(A) \neq
\infty$ for all $\condAB \in \R$ in order to exclude pathological cases.} that $\kappa^*(A) \neq
\infty$ for all $\condAB \in \R$.  
\begin{enumerate}
\item $\kappa^*$ is called \emph{$\kappa$-consistent} iff $\kappa(\omega) = \infty$ implies
  $\kappa^*(\omega) = \infty$.
%
\WEG{
\item If $\kappa^*$ is $\kappa$-consistent, then the \emph{relative change function}\index{relative change function}
  $(\kappa^* - \kappa): \Omega \to \integers \cup \{\infty\}$ is defined by
\[
(\kappa^* - \kappa) (\omega) = \left\{
\begin{array}{l@{\quad \mbox{if} \quad}l}
\kappa^*(\omega) - \kappa(\omega) & \kappa(\omega) \neq \infty\\
\infty                          & \kappa(\omega) = \infty
\end{array}
\right.
\]
}
\item  $\kappa^*$  is \emph{indifferent with
  respect to $\R$ and $\kappa$} iff $\kappa^*$ is $\kappa$-consistent and  the following two conditions hold:
\begin{enumerate}
\item[(i)] If $\kappa^*(\omega) = \infty$ then $\kappa(\omega) = \infty$, or there is $\condAB \in \R$ such that
  $\sigmaAB(\omega) \neq 1$ and $\kappa^*(\omega') = \infty$ for all $\omega'$ with
  $\sigmaAB(\omega') = \sigmaAB(\omega)$. 
\item[(ii)] $(\kappa^* - \kappa)(\omegagroup_1) = (\kappa^* - \kappa)(\omegagroup_2) \quad \mbox{whenever } \;
\sigmaR(\omegagroup_1) = \sigmaR(\omegagroup_2)$ and $\omegagroup_1 \equivo \omegagroup_2$
for  $\omegagroup_1, \omegagroup_2 \in
\Omegagroupplus^*$\index{$\Omegagroupplus^*$}, where $\Omegagroupplus^* =
\left\langle \omega \in \Omega \mid \kappa^*(\omega) \neq \infty \right\rangle$.
\index{conditional indifference!strict conditional indifference with respect
  to $\R$ and $V$}
\index{conditional indifference!conditional indifference with respect to
    $\R$ and $V$}
\end{enumerate} 
\end{enumerate} 
\end{definition}

\noindent
The principle of conditional preservation is now realized as an indifference
property:

\begin{definition}
\label{revepst_def_principle_conditional_preservation2}
Let $\kappa$ be an OCF, 
and let $\R$ be a
finite set of  conditionals.  
A revision $\kappa^*  = \kappa * \R$ satisfies the \emph{principle
    of conditional preservation} iff $\kappa^*$  is  indifferent with respect
to $\R$ and $\kappa$. 
\index{principle of conditional preservation}
\end{definition}

\WEG{
\begin{remark}
Though  the relative change function $(V^* / V)$ is not a conditional valuation
function, it may nevertheless be extended to a homomorphism $(V^* / V) :
\Omegagroupplus^* \to (\cA,\odot)$ (see Section
\ref{cond_sec_conditional_indifference}). 
Therefore, Definition
\ref{revepst_def_principle_conditional_preservation}(3) is an appropriate
modification of Definition \ref{cond_def_conditional_indifference} for
revisions.

Note that the
principle of conditional preservation is based only on observing conditional
structures, without using any acceptance conditions or taking quantifications of conditionals into account.
\end{remark}
}
So $\kappa * \R$ satisfies the principle of conditional preservation if any
change in plausibility is clearly and unambigously induced by $\R$. 
The next theorem characterizes revisions of ordinal conditional functions that
satisfy the principle of conditional preservation. The theorem is obvious by
observing  Theorem
\ref{cond_cor_char_weak_indifference_ocf}.
\begin{theorem}
\label{revepst_theorem_ocf_principle_condpres}
Let $\kappa, \kappa^*$  be OCF's, and let $\R = \{\condABi{1},$ $\ldots,
\condABi{n}\}$ be a (finite) set of conditionals in $\condL$. 
A revision $\kappa^* = \kappa * \R$ satisfies the principle of conditional
preservation  iff
$\kappa^*(A_i) \neq \infty$ for all $i, 1 \leq i \leq n$, and  there
are  numbers $\kappa_0, \kappaiplus{i}, \kappaiminus{i} \in
\rationals, 1 \leq i
\leq n$,  such that for all $\omega \in \Omega$,
\begin{equation}
\label{revepst_eq_char_indifference_ocf}
\kappa^*(\omega) = \kappa(\omega) + \kappa_0 + \sum_{1 \leq i \leq n \atop \omega \models A_i B_i}
\kappaiplus{i} + \sum_{1 \leq i \leq n \atop \omega \models A_i \ol{B_i}}
\kappaiminus{i} 
\end{equation}
\end{theorem}

Comparing Theorems \ref{cond_cor_char_weak_indifference_ocf} and
\ref{revepst_theorem_ocf_principle_condpres} with one another, we see that an
OCF $\kappa$ is indifferent with respect to a finite
set of conditionals $\R$ iff it can be taken as a revision $\kappa_0 * \R$
satisfying the principle of conditional preservation, where $\kappa_0(\omega)
= 0$ for all $\omega \in \Omega$ is the \emph{uniform ordinal conditional
  function}. 

Up to now, we have not yet taken the \emph{success condition} $\kappa^* \models \R$
into regard, postulating that the revised OCF in fact represents the
conditionals in $\R$. 
\begin{definition}
Let $\kappa,\kappa^*$ be OCF's, and let $\R$ be a set of
 conditionals. 
$\kappa^* = \kappa * \R$ is called a \emph{c-revision} iff $\kappa^* \models
\R$ and $\kappa^*$ satisfies the principle of conditional preservation.   
$\kappa$ is called a \emph{c-representation} of $\R$, iff $\kappa \models \R$
  and $\kappa$ is indifferent with respect to $\R$.   
\end{definition}
Theorems \ref{cond_cor_char_weak_indifference_ocf} and
\ref{revepst_theorem_ocf_principle_condpres} provide  simple
schemes to construct c-revisions and c-represent\-ations. The numbers $\kappa_0, \kappaiplus{i}, \kappaiminus{i} \in
\rationals, 1 \leq i \leq n$, then have to be chosen appropriately to ensure
that $\kappa^{(*)}$ is an ordinal conditional function, and such that
$\kappa^{(*)}(AB) < \kappa^{(*)}(\AnotB)$ for all conditionals $\condAB \in \R$ (cf.\
Lemma \ref{revepst_lemma_conditional_equivalences}). In the special case that
$\kappa$ is a representation of $\R$, we obtain the following corollary by some easy calculations:
\WEG{
Combining Theorem \ref{revepst_theorem_ocf_principle_condpres} with Lemma
\ref{revepst_lemma_conditional_equivalences}, we obtain
}
\begin{corollary}
\label{revepst_cor_ocf_principle_condpres}
Let $\R = \{\condABi{1}, \ldots,
\condABi{n}\}$ be a (finite) set of conditionals in $\condL$, and let $\kappa$
be an OCF.

$\kappa$ is a c-representation of $\R$  
iff
$\kappa(A_i) \neq \infty$ for all $i, 1 \leq i \leq n$, and  there
are  numbers $\kappa_0, \kappaiplus{i}, \kappaiminus{i} \in
\rationals, 1 \leq i
\leq n$,  such that for all $\omega \in \Omega$,
\begin{equation}
\label{revepst_eq_char_indifference_ocf_representation}
\kappa(\omega) = \kappa_0 + \sum_{1 \leq i \leq n \atop \omega \models A_i B_i}
\kappaiplus{i} + \sum_{1 \leq i \leq n \atop \omega \models A_i \ol{B_i}}
\kappaiminus{i}, 
\end{equation}
and
\begin{eqnarray}
\label{eq_kappaiplus_kappaiminus}
\kappaiminus{i} - \kappaiplus{i}   & > & \min_{\omega \models A_i B_i}
  \left(\sum_{j \neq i \atop \omega \models A_j B_j} \kappaiplus{j} +
        \sum_{j \neq i \atop \omega \models A_j \notB_j}
  \kappaiminus{j}\right) \\
& - &  \nonumber
\min_{\omega \models A_i \notB_i}
  \left(\sum_{j \neq i \atop \omega \models A_j B_j} \kappaiplus{j} +
        \sum_{j \neq i \atop \omega \models A_j \notB_j}
  \kappaiminus{j}\right) 
\end{eqnarray}
\end{corollary}
\begin{proof}
$\kappa$ is a c-representation of $R$ iff $\kappa \models \R$ and $\kappa$ is
indifferent with respect to $\R$. From Theorem
\ref{cond_cor_char_weak_indifference_ocf}, we obtain representation
(\ref{revepst_eq_char_indifference_ocf_representation}). Due to Lemma
\ref{revepst_lemma_conditional_equivalences}, $\kappa \models \R$ iff
$\kappa(A_i B_i) < \kappa(A_i \notB_i)$, i.e.\ iff
\begin{eqnarray*}
&& \min_{\omega \models A_i B_i} \kappa_0 + \sum_{1 \leq j \leq n \atop \omega
  \models A_j B_j} \kappaiplus{j} + \sum_{1 \leq j \leq n \atop \omega
  \models A_j \notB_j} \kappaiminus{j} \\
&&\hem < \min_{\omega \models A_i \notB_i} \kappa_0 + \sum_{1 \leq j \leq n \atop \omega
  \models A_j B_j} \kappaiplus{j} + \sum_{1 \leq j \leq n \atop \omega
  \models A_j \notB_j} \kappaiminus{j},  
\end{eqnarray*}
which is equivalent to
\begin{eqnarray*}
&& \min_{\omega \models A_i B_i} \kappaiplus{i} + \sum_{j \neq i \atop \omega
  \models A_j B_j} \kappaiplus{j} + \sum_{j \neq i \atop \omega
  \models A_j \notB_j} \kappaiminus{j} \\
&&\hem < \min_{\omega \models A_i \notB_i} \kappaiminus{i} + \sum_{j \neq i \atop \omega
  \models A_j B_j} \kappaiplus{j} + \sum_{j \neq i \atop \omega
  \models A_j \notB_j} \kappaiminus{j}.  
\end{eqnarray*}
This shows (\ref{eq_kappaiplus_kappaiminus}).
\end{proof}
The difference $\kappaiminus{i} - \kappaiplus{i}$, or the right hand side of
(\ref{eq_kappaiplus_kappaiminus}), respectively, measures the effort needed to
establish the \mbox{i-th} conditional. 
To calculate suitable constants $\kappaiplus{i},
\kappaiminus{i}$, we apply the following heuristics: To establish conditional
beliefs, one can make confirming worlds  more plausible (if required, which amounts to
choose $\kappaiplus{i} \leq 0$), or refuting worlds less plausible (if
required, which means
$\kappaiminus{i} \geq 0$). The normalizing constant $\kappa_0$ then has to be chosen
appropriately to ensure that actually an OCF is obtained. 

We prefer the second alternative, presupposing 
\begin{equation}
\label{eq_heuristics}
\kappaiminus{i} \geq 0, \mbox{ and } \kappaiplus{i} =0 \; \mbox{for} \; 1 \leq
i \leq n
\end{equation}
Then
(\ref{eq_kappaiplus_kappaiminus}) reduces to
\begin{equation}
\label{eq_reduced_kappaiminus}
\kappaiminus{i} > \min_{\omega \models A_i B_i} \sum_{j \neq i \atop \omega \models A_j \notB_j}
  \kappaiminus{j} -
\min_{\omega \models A_i \notB_i} \sum_{j \neq i \atop \omega \models A_j \notB_j}
  \kappaiminus{j}
\end{equation}
for $1 \leq i \leq n$.
If there are worlds $\omega$ with neutral conditional structure,
$\sigmaR(\omega) = 1$, we may set $\kappa_0 = 0$. So, we obtain a c-representation
of $\R$ via
\begin{equation}
\label{eq_kappa_reduced}
\kappa(\omega) = \sum_{1 \leq i \leq n \atop \omega \models A_i \notB_i} \kappaiminus{i},
\end{equation} 
where the $\kappaiminus{i}$ have to satisfy (\ref{eq_reduced_kappaiminus}).
\begin{example}
\label{revepst_ex_penguin2}
We will use Corollary \ref{revepst_cor_ocf_principle_condpres} and the
heuristics (\ref{eq_heuristics}) to obtain a
c-representation (\ref{eq_kappa_reduced}) of the conditionals $\R = \{r_1, \ldots, r_5\}$ of Example
\ref{revepst_ex_penguin}. 
To calculate constants $\kappaiminus{i}$ according to (\ref{eq_reduced_kappaiminus}),  Table \ref{revepst_figure_penguin} proves to be helpful. We only have to focus on the
$\mathbf{a}^-$-labels of worlds, and we obtain
\begin{eqnarray*}
&&\kappaiminus{5}, \kappaiminus{4}, \kappaiminus{1} > 0,\\
&&\kappaiminus{3} > \min\{\kappaiminus{1},\kappaiminus{2}\}, \;
\kappaiminus{2} > \min\{\kappaiminus{1},\kappaiminus{3}\}
\end{eqnarray*} 
So we set 
\begin{equation}
\label{eq_penguin_kappaiminus}
\kappaiminus{5} = \kappaiminus{4} = \kappaiminus{1} = 1, \;
\kappaiminus{2} = \kappaiminus{3} = 2. 
\end{equation}
Since there are also worlds $\omega$ with $\sigmaR(\omega) = 1$ (cf.\ Table
\ref{revepst_figure_penguin}), we set $\kappa_0 = 0$. So we obtain a c-representation
of $r_1, \ldots, r_5$ by
\begin{equation}
\label{eq_penguin_kappa}
\kappa(\omega) = \sum_{1 \leq i \leq 5 \atop \omega \models A_i \notB_i} \kappaiminus{i},
\end{equation} 
where the $A_i,B_i$'s are the antecedents and consequents of the rules $r_i$
and the $\kappaiminus{i}$ are defined as in (\ref{eq_penguin_kappaiminus}) for $1
\leq i \leq 5$ (see also Table \ref{revepst_figure_systemz} in Example \ref{revepst_ex_systemz} below).
\end{example}

\section{A comparison with system-Z and system-$\Zstar$}
\label{sec_comparison_System_Z}
A well-known method to represent a (finite) set $\R= \{ r_i = \condABi{i} \mid
1 \leq i \leq n\}$ of conditionals by an OCF  is to apply the \emph{system-Z}\index{system-Z} of
Goldszmidt and Pearl \cite{GoldszmidtPearl92,GoldszmidtPearl96}. The
corresponding ranking function $\kappa^z$ is given by
\begin{equation}
\label{revepst_eq_systemz}
\kappa^z(\omega) = \left\{
\begin{array}{l}
0, \; \mbox{if $\omega$ does not falsify any $r_i$},\\ 
1 + \max\limits_{1 \leq i \leq n \atop \omega \models A_i \notB_i} Z(r_i), \mbox{otherwise}
\end{array}
\right.
\end{equation}
where $Z$ is an ordering on  $\R$ observing the (logical)
interactions of the conditionals (for a detailed description of $Z$, see, for
instance, \cite{GoldszmidtPearl96}). $\kappa^z$ assigns to each world $\omega$
the lowest possible rank admissible with respect to the constraints in
$\R$. Comparing (\ref{revepst_eq_systemz}) with
(\ref{cond_eq_char_weak_indifference_ocf}), we see that in general, $\kappa^z$
is \emph{not} a c-representation of $\R$, since in its definition (\ref{revepst_eq_systemz}),
\emph{maximum} is used instead of \emph{summation} (see Example
\ref{revepst_ex_systemz} below). The numbers $Z(r_i)$,
however, may well serve to define appropriate constants $\kappaiminus{i}$ in
(\ref{cond_eq_char_weak_indifference_ocf}). Setting $\kappa_0 =
\kappaiplus{i}  = 0$, and $\kappaiminus{i} = Z(r_i) +1$ for $1 \leq i \leq
n$, we obtain from $Z$ a
c-representation $\kappa_c^z$ of $\R$ via 
\begin{equation}
\label{revepst_eq_kappazc}
\kappa^z_c(\omega) =\left\{
\begin{array}{l}
0, \; \mbox{if $\omega$ does not falsify any $r_i$},\\ 
 \sum\limits_{1 \leq i \leq n \atop \omega \models A_i \notB_i}
(Z(r_i) +1), \quad
\mbox{otherwise}.
\end{array}
\right.
\end{equation}

An even more sophisticated representation is obtained by combining the
system-Z approach with the principle of maximum entropy (\emph{ME-principle}), yielding system-$\Zstar$
\cite{GoldszmidtMorrisPearl93}. The corresponding $\Zstar$-rankings of the
conditionals in $\R$ have to satisfy the following equation (see equation (16)
in \cite[p.\ 225]{GoldszmidtMorrisPearl93})
\begin{eqnarray}
\label{eq_Zstar}
&& \Zstar(r_i) + \min_{\omega \models A_i \notB_i} \sum_{j \neq i \atop \omega
  \models A_j \notB_j} \Zstar(r_j) \\
&& \hem = 1 + \min_{\omega \models A_i B_i} \sum_{j \neq i \atop \omega \models A_j
  \notB_j} \Zstar(r_j), \nonumber
\end{eqnarray}
and $\kappastar$ is then calculated by
\begin{equation}
\label{eq_kappastar}
\kappastar(\omega) = \sum_{\omega \models A_i \notB_i} \Zstar(r_i)
\end{equation}
(see equation (18) in \cite[p.\ 225]{GoldszmidtMorrisPearl93}). For so-called
\emph{minimal-core sets} -- these are sets $\R$ allowing each conditional to
be separable from the other rules by restricting conditional interactions --,
a procedure is given to calculate $\Zstar$-rankings in
\cite{GoldszmidtMorrisPearl93}.  

Like our method, system-$\Zstar$ makes use of summation instead of
maximization, as in system-Z. And  equations (\ref{eq_Zstar}) determining the
$\Zstar$-rankings look similar to our inequality constraints 
(\ref{eq_kappaiplus_kappaiminus}). More exactly, if we follow the heuristics
(\ref{eq_heuristics}) and set $\Zstar(r_i) =
 \kappaiminus{i}$, then system-$\Zstar$ turns out to be a special instance of
 our more general scheme in Corollary \ref{revepst_cor_ocf_principle_condpres}. In particular, system-$\Zstar$ yields a c-representation. 

This similarity is not accidental -- the ME-principle not
 only provides a powerful base for system-$\Zstar$, but also influenced the idea
 of conditional indifference presented in this paper. In
 \cite{Kern-Isberner97e}, we characterized the ME-principle by four axioms,
 one of which was the \emph{postulate of conditional
   preservation}. Conditional preservation for probability functions there was
 realized in full analogy to that for OCF's defined here. So both c-representations
 and  ME-distributions comply with a fundamental principle for representing 
 conditionals, and it is this principle of conditional preservation (or
 principle of conditional indifference, respectively) that is responsible for
 a peculiar thoroughness and accuracy when incorporating conditionals. Since we
 realized this principle completely in a semi-quantitative
 setting, we did not have to refer to probabilities and to ME-distributions, and
 we were able to formalize the acceptance conditions,
 (\ref{eq_kappaiplus_kappaiminus}), in a purely qualitative manner.  

We will illustrate our method by various examples which are taken from \cite{GoldszmidtMorrisPearl93}
and \cite{GoldszmidtPearl96} to allow a direct comparison with system-Z and
system-$\Zstar$. 

%
\begin{example}
\label{revepst_ex_systemz}
Consider once again the conditionals $r_1, \ldots, r_5$ from Example
\ref{revepst_ex_penguin}. 
Here we have $Z(r_1) = Z(r_4) = Z(r_5) = 0$ and $Z(r_2) = Z(r_3) = 1$ (for the
details, see \cite[p.\ 69]{GoldszmidtPearl96}). By setting $\kappaiminus{i} =
Z(r_i) + 1$ for \mbox{$1\leq i \leq 5$}, we obtain the same constants,
(\ref{eq_penguin_kappaiminus}), as in Example \ref{revepst_ex_penguin2}. 
Furthermore, by applying the procedure \emph{Z-rank} in
\cite{GoldszmidtMorrisPearl93}, we calculate $\Zstar(r_1) = \Zstar(r_4) =
\Zstar(r_5) = 1$ and $\Zstar(r_2) = \Zstar(r_3) = 2$, therefore also $\Zstar(r_i) =
\kappaiminus{i}$, \mbox{$1\leq i \leq 5$}. 
 So, in
this example, 
$\kappa^z_c$ from (\ref{revepst_eq_kappazc}) actually is the OCF from (\ref{eq_penguin_kappa}) and coincides
with $\kappastar$. For instance, 
\begin{eqnarray*}
\kappa^z_c(\ol{p}\ol{b}\,\ol{f}wa) &=& 0,\\ 
\kappa^z_c(pb \ol{f}wa) &=& \kappaiminus{1} = 1,\\ 
\kappa^z_c(p\ol{b}fw\ol{a}) &=& \kappaiminus{2} + \kappaiminus{3} + \kappaiminus{5} = 5.
\end{eqnarray*}
\begin{table}[h]
\begin{displaymath}
\begin{array}{|l|c|c||l|c|c|}
\hline
\omega & \kappa^z(\omega) & \kappa^z_c(\omega) & \omega & \kappa^z(\omega) &
\kappa^z_c(\omega) \\
\hline
\rule{0mm}{4mm} pbfwa & 2 & 2 & \ol{p}bfwa & 0 & 0 \\
pbfw\ol{a} & 2 & 3 & \ol{p}bfw\ol{a} & 1 & 1 \\
pbf\ol{w}a & 2 & 3 & \ol{p}bf\ol{w}a & 1 & 1 \\
pbf\ol{w}\,\ol{a} & 2 & 4 & \ol{p}bf\ol{w}\,\ol{a} & 1 & 2 \\

\rule{0mm}{4mm} pb\ol{f}wa & 1 & 1 & \ol{p}b\ol{f}wa & 1 & 1 \\
pb\ol{f}w\ol{a} & 1 & 1 & \ol{p}b\ol{f}w\ol{a} & 1 & 1 \\
pb\ol{f}\ol{w}a & 1 & 2 & \ol{p}b\ol{f}\ol{w}a & 1 & 2 \\
pb\ol{f}\ol{w}\,\ol{a} & 1 & 2 & \ol{p}b\ol{f}\ol{w}\,\ol{a} & 1 & 2 \\

\rule{0mm}{4mm} p\ol{b}fwa & 2 & 4 & \ol{p}\ol{b}fwa & 0 & 0 \\
p\ol{b}fw\ol{a} & 2 & 5 & \ol{p}\ol{b}fw\ol{a} & 1 & 1 \\
p\ol{b}f\ol{w}a & 2 & 4 & \ol{p}\ol{b}f\ol{w}a & 0 & 0 \\
p\ol{b}f\ol{w}\,\ol{a} & 2 & 5 & \ol{p}\ol{b}f\ol{w}\,\ol{a} & 1 & 1 \\

\rule{0mm}{4mm} p\ol{b}\,\ol{f}wa & 2 & 2 & \ol{p}\ol{b}\,\ol{f}wa & 0 & 0 \\
p\ol{b}\,\ol{f}w\ol{a} & 2 & 2 & \ol{p}\ol{b}\,\ol{f}w\ol{a} & 0 & 0 \\
p\ol{b}\,\ol{f}\ol{w}a & 2 & 2 & \ol{p}\ol{b}\,\ol{f}\ol{w}a & 0 & 0 \\
p\ol{b}\,\ol{f}\ol{w}\,\ol{a} & 2 & 2 & \ol{p}\ol{b}\,\ol{f}\ol{w}\,\ol{a} & 0
& 0 \\
\hline
\end{array}
\end{displaymath}
\caption{\label{revepst_figure_systemz} Rankings for Example \ref{revepst_ex_systemz}}
\end{table}
%
In  Table \ref{revepst_figure_systemz}, 
we list the ranks of all possible worlds, first 
computed by system-Z, according to (\ref{revepst_eq_systemz}), and then 
computed as a c-representation, $\kappa^z_c$, of $\R$, according to
(\ref{revepst_eq_kappazc}). 

This table  
 reveals clearly that
$\kappa^z$ is not a c-representation of $\R$: Associating  symbols
$\aiplus{i}, \aiminus{i}$ with the conditionals $r_i$ in $\R$, \mbox{$1 \leq i
  \leq 5$}, respectively, we obtain 
\[
\sigmaR\left(\D\frac{pbfwa \cdot \ol{p}bfw\ol{a}}{pbfw\ol{a} \cdot
  \ol{p}bfwa}\right) = \D\frac{\aiplus{1}\aiplus{2} \aiminus{3} \aiplus{4} \aiplus{5} \cdot \aiplus{1} \aiplus{4}
  \aiminus{5}}{\aiplus{1} \aiplus{2} \aiminus{3} \aiplus{4} \aiminus{5} \cdot \aiplus{1} \aiplus{4}
  \aiplus{5}}
 = 1,
\]
but
$\kappa^z\left(\D\frac{pbfwa \cdot \ol{p}bfw\ol{a}}{pbfw\ol{a} \cdot
  \ol{p}bfwa}\right)$ $= \kappa^z(pbfwa) + \kappa^z(\ol{p}bfw\ol{a}) -
  \kappa^z(pbfw\ol{a}) - \kappa^z(\ol{p}bfwa)= 2 + 1 -2  = 1 \neq 0$.

What is the actual benefit of this formal principle of conditional preservation?
Comparing $\kappa^z(\omega)$ to $\kappa^z_c(\omega)$, we see that $\kappa^z_c$
is more fine-grained. Therefore, it represents more conditionals. Consider,
e.g., the conditional $(w | pb\ol{f}a)$ -- does a non-flying, but airborne
penguin possess wings or not? $\kappa^z$ does not know, we have $\kappa^z(pb\ol{f}aw)
= \kappa^z(pb\ol{f}a\ol{w}) = 1$. On the other hand, $\kappa^z_c$ accepts this
conditional: $\kappa^z_c(pb\ol{f}aw) = 1 < 2 =
\kappa^z_c(pb\ol{f}a\ol{w})$. To show that this is more than pure speculation,
consider the conditional structures of $pb\ol{f}aw$ and $pb\ol{f}a\ol{w}$:
$\sigmaR(pb\ol{f}aw) = \aiminus{1} \aiplus{2} \aiplus{3} \aiplus{4}$,
$\sigmaR(pb\ol{f}a\ol{w}) = \aiminus{1} \aiplus{2} \aiplus{3}
\aiminus{4}$. Thus, except for $r_4$, both worlds behave exactly the same with
respect to the conditionals in $\R$, but, due to $r_4$, our penguin is
supposed to have wings since it is a bird. 

Similar arguments apply when considering the conditionals $(a | pbf), (a |
pbfw), (w | pbf), (w | pbfa)$: $\kappa^z$ is totally indifferent when
confronted with flying super-penguins - it assigns the same degree of
plausibility, 2, to any of the involved worlds. So, it accepts neither of
these conditionals, whereas $\kappa^z_c$ accepts all of them.  

Finally, let us  consider the conditional $(a | bfw)$. It is accepted by
$\kappa^z$, as well as by $\kappa^z_c$, as may easily be checked. But what
happens when the variable $p$ is taken into account? $\kappa^z$ establishes
$(a | \ol{p}bfw)$, but is undecided with respect to $(a | pbfw)$. By contrast,
$\kappa^z_c$ not only accepts both of these conditionals, but also establishes
them with equal strength: 
\begin{equation}
\label{eq_example_penguin}
\kappa^z_c\left(\frac{pbfwa}{pbfw\ol{a}}\right) =
  \kappa^z_c\left(\frac{\ol{p}bfwa}{\ol{p}bfw\ol{a}}\right) = -1.
\end{equation} 
This is a simple consequence of the principle of conditional preservation in
  this case, since $\sigmaR\left(\D\frac{pbfwa \cdot \ol{p}bfw\ol{a}}{pbfw\ol{a} \cdot
  \ol{p}bfwa}\right)=1$ (see above). (\ref{eq_example_penguin}) is justified because each of the
  involved quotients has the same conditional structure, i.e.\ shows the same
  behavior, with respect to $\R$. By considering arbitrary group elements in
  $\Kern \sigmaR$, even very complicated interrelationships between
  degrees of strength associated with conditionals can be observed. 
\end{example}

Therefore, using System-Z and the $\max$-operator means to establish conditionals
 only on a superficial level, whereas obeying the principle of
conditional preservation ensures that conditional knowledge is propagated
 thoroughly and deeply in plausibility structures of epistemic states. This
 well-behavedness with respect to subconditionals is also observed in
 \cite{GoldszmidtMorrisPearl93}. It is illustrated by the next example, too:

\begin{example}
Consider the conditionals
\[
\begin{array}{l@{\; : \;}l@{\quad}l}
r_1 & (f | s) & \mbox{\emph{Swedes are fair-haired.}}\\
r_2 & (t | s) & \mbox{\emph{Swedes are tall.}}
\end{array}
\]
We apply Corollary \ref{revepst_cor_ocf_principle_condpres} and heuristics
(\ref{eq_heuristics}) and calculate $\kappaiminus{1}, \kappaiminus{2} \geq 0$.
So we set $\kappaiminus{1} =  \kappaiminus{2}= 1$, and we obtain a
c-representation, $\kappa$, of the form (\ref{eq_kappa_reduced}). Then not
only the subconditional $(f |st)$ of $(f | s)$ is accepted, but also the
subconditional $(f | s\ol{t})$, because $\kappa(sf\ol{t}) = 1 < 2 =
\kappa(s\ol{f}\,\ol{t})$. 

Goldszmidt, Morris and Pearl \cite{GoldszmidtMorrisPearl93} compared this
situation to the one where instead of $r_1,r_2$,  merely the conditional $(ft|s)$ is learned. Here,
only the first subconditional, $(f|st)$, is accepted, but not the second one, $(f | s\ol{t})$. This becomes
intelligible by considering conditional structures: $sf\ol{t}$ and
$s\ol{f}\,\ol{t}$ both show the same behavior with respect to $(ft|s)$
(namely, they refute it), while $sft$ and $s\ol{f}t$ show different
behaviors. 

Observing conditional structures emphasizes once again that in general, the joint integration of
 conditionals cannot be achieved by learning only one conditional --
 conditionals resist to propositional treatment. In our
 framework, each
 conditional constitutes an independent piece of knowledge. 
\end{example}
 
Finally, let us consider an example that cannot be dealt with by
system-$\Zstar$ in a straightforward manner because the involved set of
conditionals is not a minimal-core set:

\begin{example}
\label{ex_non_mc_set}
Let $\R$ consist of the rules
\[
r_1 \; : \; (b|a), \quad r_2 \; : \; (c|b), \quad r_3 \; : \; (c|a).
\]
We list the conditional structures in Table \ref{table_non_mc_set} to make
argumentation easier.
\begin{table}[h]
\begin{displaymath}
\begin{array}{|l|l||l|l|}
\hline
\omega & \sigmaR(\omega) & \omega & \sigmaR(\omega) \\ 
\hline
\rule{0mm}{4mm} abc & \aiplus{1} \aiplus{2} \aiplus{3} & \ol{a}bc &
\aiplus{2}\\
ab\ol{c} & \aiplus{1} \aiminus{2} \aiminus{3} & \ol{a}b\ol{c} & \aiminus{2}\\
a\ol{b}c & \aiminus{1} \aiplus{3} & \ol{a} \ol{b} c & 1 \\
a\ol{b}\ol{c} & \aiminus{1}\aiminus{3} & \ol{a} \ol{b} \ol{c} & 1 \\
\hline
\end{array}
\end{displaymath}
\caption{\label{table_non_mc_set} Conditional structures for Example \ref{ex_non_mc_set}}
\end{table}
$\R$ is not a minimal-core set in the sense of \cite{GoldszmidtMorrisPearl93}
because $r_3$ is only refuted by worlds that also refute either $r_1$ or
$r_2$. So equation (\ref{eq_Zstar}) is not solvable to yield
$\Zstar$-rankings. 

For our approach, however, dealing with $\R$ is no problem: Using
(\ref{eq_heuristics}) and (\ref{eq_reduced_kappaiminus}), we obtain
\[
\kappaiminus{1}, \kappaiminus{2} > 0, \quad \kappaiminus{3} > 0 -
\min\{\kappaiminus{1}, \kappaiminus{2}\}
\]
We set $\kappaiminus{1} = \kappaiminus{2} = 1$ and $\kappaiminus{3} =
0$, and we obtain by (\ref{eq_kappa_reduced}) an appropriate ranking function,
$\kappa$ (see Table \ref{table_non_mc_set2}).
\begin{table}[h]
\begin{displaymath}
\begin{array}{|l|l|l||l|l|l|}
\hline
\omega & \kappa(\omega) & \kappa_1(\omega) & \omega & \kappa(\omega) &
\kappa_1(\omega) \\ 
\hline
\rule{0mm}{4mm} abc & 0 & 0 & \ol{a}bc & 0 & 0 \\
ab\ol{c} & 1 & 2 & \ol{a}b\ol{c} & 1 & 1 \\
a\ol{b}c & 1 & 1 & \ol{a} \ol{b} c & 0 & 0 \\
a\ol{b}\ol{c} & 1 & 2 & \ol{a} \ol{b} \ol{c} & 0 & 0 \\
\hline
\end{array}
\end{displaymath}
\caption{\label{table_non_mc_set2} Rankings for Example \ref{ex_non_mc_set}}
\end{table}
Actually, $r_3$ seems to be redundant since $\kappaiplus{3} = \kappaiminus{3}
= 0$, and it is in fact already established by $r_1$ and $r_2$. But what about
the subconditionals $(c |ab)$ and $(c | a\ol{b})$ of $r_3$? While the first
one is accepted, due to the impact of $r_2$, the second subconditional is not,
since $r_3$ is taken to be redundant 
(see the conditional structures in
Table \ref{table_non_mc_set}). 

To ensure the thorough propagation of conditional knowledge to
subconditionals, we have to postulate $\kappaiminus{i} > 0$ in
(\ref{eq_heuristics}), in order to protect the influence of each conditional
against numerical cancellations. With $\kappaiminus{1} = \kappaiminus{2} =
\kappaiminus{3} = 1$, we obtain the OCF $\kappa_1$ in Table
\ref{table_non_mc_set2}, which accepts both subconditionals of $r_3$, as
desired. 
\end{example}

\section{Unifying qualitative and quantitative approaches}
\label{sec_unifying}
We defined the
principle of conditional preservation as an indifference property of the
revised ranking function (cf.\ Definition
\ref{revepst_def_principle_conditional_preservation2}) -- conditional
preservation means to maintain numerical relationships. Therefore, it appears
here as an essentially quantitative notion. Nevertheless, it is applied to ranking
functions which are used in a qualitative, or at least,
semi-quantitative setting. Moreover, note that conditional
indifference is based on the notion of conditional structures which are
represented in a completely symbolic way. So, what is the connection between the
principle of conditional preservation introduced here, and the approaches to
conditional preservation in a purely qualitative framework, as were
proposed in \cite{DarwichePearl97} and in \cite{Kern-Isberner99b}?

In \cite{Kern-Isberner99b}, we advanced a set of postulates apt to guide the
revisions of epistemic states, $\Psi$, by conditional beliefs, $\condAB$. There, the idea of conditional preservation was put in formal
terms by making use of two relations, \emph{subconditionality}, $\sqsubseteq$ (see (\ref{eq_subconditional})), and
\emph{perpendicularity}, $\perp$, on the set of conditionals:
\[
(D | C) \perp (B | A) \; \mbox{iff either } C \models AB, \mbox{ or } \AnotB,
\mbox{ or } \notA.
\]
If $(D | C) \perp (B | A)$, then for all worlds which $\condCD$ may be applied
to, $\condAB$ has the same effect and thus, it yields no further
partitioning -- $\condAB$ is \emph{irrelevant} for $\condCD$. 

Conditional preservation then was described in \cite{Kern-Isberner99b} by the
following three postulates: 
\begin{description}
\item[{\bf (CR5)}] If $\condCD \perp \condAB$ then $\Psi \models \condCD$ iff
  \\ $\Psi * \condAB \models \condCD$.
\item[{\bf (CR6)}] If $\condCD \sqsubseteq \condAB$ and $\Psi \models \condCD$
  then \\ $\Psi * \condAB \models \condCD$.
\item[{\bf (CR7)}] If $\condCD \sqsubseteq (\notB | A)$ and $\Psi *
  \condAB \models \condCD$ then $\Psi \models \condCD$.
\end{description}
These axioms cover the postulates of Darwiche and Pearl in \cite{DarwichePearl97} (see \cite{Kern-Isberner99b,Kern-Isberner99a}) and support
their intuitive ideas with more formal arguments. 
Applying Definition
\ref{revepst_def_principle_conditional_preservation2} 
 to the case $\R
= \{\condAB\}$, the principle of conditional preservation reduces to postulate
\[
(\kappa^* - \kappa)(\omega_1) = (\kappa^* - \kappa)(\omega_2) \quad
\mbox{if} \quad \condAB(\omega_1) = \condAB(\omega_2)
\]
for the revised function $\kappa^* = \kappa * \R$ (cf.\ also
\cite{Kern-Isberner99c}). It can then be shown that such a revision also
satisfies the postulates (CR5)-(CR7) stated above \cite{Kern-Isberner99c}. 

This means that the principle of conditional preservation, phrased in its full
(numerical) complexity in this paper, also covers the approaches to
conditional preservation proposed in a qualitative framework.

To guarantee the thorough propagation of conditional knowledge to
subconditionals, we may supplement (CR5)-(CR7) with another postulate:
\begin{description}
\item[{\bf (CR8)}] If $\condCD \sqsubseteq \condAB$ and $\Psi \not\models
  (\ol{D} | C)$, then \\ $\Psi * \condAB \models \condCD$.
\end{description}
(CR8) clearly exceeds the paradigm of conditional preservation, in favor of
imposing conditional structure as long as there are no conflicts.

Axioms (CR5)-(CR8) only deal, however, with revising an epistemic state by
one single conditional. A topic of our ongoing research is to generalize them
so as to apply for incorporating sets of conditionals, too.

\section{Summary and outlook}
\label{sec_summary}
In this paper, we presented an approach to realize the idea of conditional
preservation in knowledge representation and belief revision in a very
comprehensive way. We introduced a formal notion of conditional structure of worlds,  and
then we phrased a \emph{principle of conditional preservation} for (revisions of) ordinal
conditional functions as  indifference with respect to these
structures. We presented simple schemes to construct ordinal conditional
functions observing this principle,  and we compared our approach to  system-Z and system-$\Zstar$.   Finally, we showed that the principle formalized here 
also covers other, qualitative approaches to conditional preservation.

In \cite{Kern-Isberner99d}, we developed these ideas in an even more general
setting, with so-called \emph{conditional valuation functions} representing
epistemic states. Ordinal conditional functions are special instances of these
functions, as well as probability distributions and possibility
distributions. So the principle of conditional preservation, as was presented here,  can be applied to
 nearly all major (semi-)quantitative representation forms for epistemic
states. Actually, this paper continues and extends work begun in \cite{Kern-Isberner97e}.

The algebraic notion of  \emph{conditional structures}  of worlds played a
crucial part for conditional preservation here. Its representation via group
theory does not only provide an elegant methodological framework for handling
conditionals. In \cite{Kern-Isberner99d}, we present an approach to solve the
``inverse representation problem'' -- which conditionals are most appropriate
to represent an epistemic state? -- by making extensive use of group
theoretical means to elaborate conditional structures. 
\bigskip

\noindent
\textbf{Acknowledgements}\\
I am grateful to three anonymous referees whose comments helped me to improve
the presentation of my results.




\end{document}